\documentclass[10pt,twocolumn,letterpaper]{article}

\usepackage{cvpr}
\usepackage{times}
\usepackage{epsfig}
\usepackage{graphicx}
\usepackage{amsmath}
\usepackage{amssymb}
\usepackage{cite}

\usepackage{multirow}

\def\R{\mathbb{R}}

\newcommand{\norm}[1]{\|#1\|}

\usepackage{hyperref} 

\cvprfinalcopy 

\def\cvprPaperID{383} 

\begin{document}

\title{Robust Compressed Sensing and Sparse Coding with the Difference Map}

\author{Will Landecker\thanks{This work was supported by National Science Foundation Grant EF- 1137929, ``The Small Number Limit of Biological Information Processing,'' and the Emergent Institutions Project.}\\
Portland State University\\
{\tt\small landeckw@cs.pdx.edu}
\and
Rick Chartrand\thanks{This work was supported by the U.S. Department of Energy  through the LANL/LDRD Program, and by the University of California Laboratory Fees Research Program.}\\
Los Alamos National Laboratory\\
{\tt\small rickc@lanl.gov}
\and
Simon DeDeo$^*$\\
Santa Fe Institute\\
Indiana University \\
{\tt\small simon@santafe.edu}
}

\maketitle

\begin{abstract}
In compressed sensing, we wish to reconstruct a sparse signal $x$ from observed data $y$. In sparse coding, on the other hand, we wish to find a representation of an observed signal $y$ as a sparse linear combination, with coefficients $x$, of elements from an overcomplete dictionary. 
 While many algorithms are competitive at both problems when $x$ is very sparse, it can be challenging to recover $x$ when it is \emph{less} sparse. We present the \emph{Difference Map}, which excels at sparse recovery when sparseness is lower and noise is higher. The Difference Map out-performs the state of the art with reconstruction from random measurements and natural image reconstruction via sparse coding.


\end{abstract}


\section{Introduction} \label{sec:intro}

In compressed sensing (CS), we are given a measurement matrix $\Phi \in \R^{m\times n}$ (where $m < n$),  observed data $y \in \R^m$, and we wish to recover a \emph{sparse} $x \in \R^n$ such that
\begin{equation*}
y = \Phi x.
\end{equation*}
The compressed sensing problem can then be written as
\begin{equation}\label{eqn:np-hard}
\arg\min_{x} \norm{ x }_0 \; \text{ subject to } \; y = \Phi x,
\end{equation}
where $\norm{ \cdot }_0$ is the $\ell^0$ penalty function, giving the number of nonzero elements. In the noisy case, where 
\begin{equation*}
\tilde{y} = \Phi x + \epsilon \cdot \mathcal{N}(0,\sigma)
\end{equation*}
is assumed to be a noisy observation, we often replace the linear constraints with quadratic ones:
\begin{equation}\label{eqn:np-hard2}
\arg\min_{x} \norm{ x }_0 \; \text{ subject to } \; \norm{ \tilde{y} - \Phi x }_2^2 \le \delta 
\end{equation}
for some $\delta > 0$. The problem \eqref{eqn:np-hard2} can also be used for sparse coding (SC); in this setting, $\Phi$ is an overcomplete dictionary, $y$ is a signal (such as an image patch), and we seek a sparse coefficient vector $x$.

Recently, a variety of algorithms have achieved good results for a variety of CS and SC problems, including Least Angle Regression (LARS) \cite{efron-2004-least}, Subspace Pursuit \cite{dai2009}, Matching Pursuit and its variants \cite{donoho2012,tropp2007}, Iterative Hard Thresholding (IHT) and its variants \cite{2010Blumensath,Blumensath2009,Blumensath2012}, Iteratively Reweighted Least Squares (IRLS) \cite{Chartrand2008}, and the Alternating Direction Method of Multipliers (ADMM) \cite{Boyd2011,Chartrand2013}. 

Because solving problems \eqref{eqn:np-hard} and \eqref{eqn:np-hard2} directly is known to be NP-hard \cite{natarajan1995sparse}, some approaches relax the $\ell^0$ penalty to the convex $\ell^1$ norm \cite{Tibshirani1996, efron-2004-least}, while some attempt to address the $\ell^0$ case directly \cite{dai2009,tropp2007,2010Blumensath}, and still others consider any number of $\ell^p$ (quasi-)norms for $0 < p \le 1$ \cite{Boyd2011,Chartrand2008, Chartrand2013}.
In general, the challenge for CS and SC algorithms is to balance two competing constraints on the solution $x^*$: to accurately reconstruct the observed data $y$, and to be sparse.

This paper presents a method for solving CS and SC problems without relaxing the $\ell^0$ constraint, using a general method known as the Difference Map \cite{Elser2007}. The Difference Map (DM) has been used to solve a wide variety constraint-intersection problems. Given sets $A$ and $B$ and distance-minimizing projections $P_A$ and $P_B$\footnote{By distance-minimizing projection, we mean that $P_A(x_0) = \min_x \norm{x_0 - x}_2$ subject to $x \in A$, and likewise for $P_B$.}, respectively, DM searches for a point $x^* \in A \cap B$. One iteration of DM is defined by $x \gets D(x)$, where
\begin{equation}
D(x) = x + \beta \left[ P_A \circ f_B(x) - P_B \circ f_A(x) \right] \label{eqn:DM}
\end{equation}
where
\begin{gather*}
f_A(x) = P_A(x) - \left( P_A(x) - x  \right) / \beta \\
f_B(x) = P_B(x) + \left( P_B(x) - x \right) / \beta
\end{gather*}
and $\beta \neq 0$. One can test for convergence by monitoring the value $| P_A \circ f_B(x) - P_B \circ f_A(x) |$, which vanishes when a solution is found.
Recently, DM has achieved state-of-the-art performance on a variety of NP-hard nonconvex optimization problems including protein folding, $k$-SAT, and packing problems \cite{Elser2007}.

The rest of this paper is organized as follows. In Section~\ref{sec:dm}, we introduce an adaptation of the Difference Map for compressed sensing and sparse coding, which we compare at a high level to other well-known algorithms. In Section~\ref{sec:random-matrix}, we compare the algorithms on CS problems using random measurements, and we reconstruct natural images via SC in Section~\ref{sec:image}. 

\section{Compressed Sensing and Sparse Coding with the Difference Map}\label{sec:dm}

Given a matrix $\Phi \in \R^{m \times n}$ (where $m<n$) and data $y \in \R^m$, we wish to find a sparse vector $x^* \in \R^n$ that is a solution to problem \eqref{eqn:np-hard2}. We apply the Difference Map to this problem by defining the constraint sets
\begin{gather*}
A = \{x \in \R^n : \norm{x}_0 \le s\} \\
B = \{x \in \R^n : \norm{ \Phi x - y }_2 ^2 < \delta \}
\end{gather*}
for a pre-defined positive integer $s$ and scalar $\delta > 0$.

The minimum-distance projection onto $A$ is known as \emph{hard thresholding}, and is defined by
\begin{equation}\label{eqn:hard-thresholding}
P_A(x) = [ x ]_s, 
\end{equation}
where $[ x ]_s$ is obtained by setting to zero the $n - s$ components of $x$ having the smallest absolute values.

A minimum-distance projection onto the set $B$ involves solving a quadratically-constrained quadratic programming problem, which can be very costly. We approximate this projection with:
\begin{equation}\label{eqn:pseudo}
P_B(x) = x - \Phi^+ (\Phi x - y) 
\end{equation}
where $\Phi^+ = \Phi^T (\Phi \Phi^T)^{-1}$ is the Moore-Penrose pseudo-inverse of $\Phi$.

The motivation for \eqref{eqn:pseudo} comes from a simplification of our definition for $B$. Consider the set with linear constraints (as in \eqref{eqn:np-hard}):
\begin{equation*}
\{x \in \R^n : \Phi x = y \}.
\end{equation*}
The minimum distance projection onto this set is given by the linearly constrained quadratic programming (QP) problem
\begin{equation*}
P(x_0) = \arg\min_{x \in \R^n} \tfrac{1}{2}\norm{ x-x_0 }_2^2 \;\text{ such that }\; \Phi x = y.
\end{equation*}
The Lagrangian of this QP is
\begin{equation*}
\mathcal{L}(x,\lambda) = \tfrac{1}{2}\norm{ x - x_0 }_2^2 + \lambda( \Phi x-y).
\end{equation*}
The $x$ that solves the QP is a minimizer of $\mathcal{L}$, and is found by setting $\nabla_x(\mathcal{L})=0$, which yields
\begin{equation}\label{eqn:x-x0}
x = x_0 + \Phi^T \lambda. 
\end{equation}
Plugging \eqref{eqn:x-x0} into $y = \Phi x$ and solving for $\lambda$ gives
\begin{equation*}
\lambda = (\Phi \Phi^T)^{-1} (\Phi x_0 - y).
\end{equation*}
Finally, we plug this into \eqref{eqn:x-x0} to get
\begin{equation*}
x = x_0 - \Phi^T (\Phi \Phi^T)^{-1}(\Phi x_0 - y)
\end{equation*}
as in \eqref{eqn:pseudo}.

Although the motivation for \eqref{eqn:pseudo} comes from the assumption of non-noisy observations $y = \Phi x$, we will see that it performs very well in the noisy case.


It is worth noting that the pseudo-inverse is expensive to compute, though it only needs to be computed once. Thus in the case of sparse image reconstruction where each image patch is reconstructed independently, amortizing the cost of computing $\Phi^+$ over all image patches significantly reduces the pre-computation overhead.

%

\subsection{Comparison to Other Algorithms}

In what follows, we compare the Difference Map to a representative sample of commonly-used algorithms for solving CS and SC problems: Least Angle Regression (LARS) \cite{efron-2004-least}, Stagewise Orthogonal Matching Pursuit (StOMP) \cite{donoho2012}, Accelerated Iterative Hard Thresholding (AIHT) \cite{Blumensath2012}, Subspace Pursuit \cite{dai2009}, Iteratively-Reweighted Least Squares (IRLS) \cite{Chartrand2008} and Alternating Direction Method of Multipliers (ADMM) \cite{Boyd2011,Chartrand2013}. As a final point of comparison, we test the Alternating Map (AM) defined by $x \gets P_A(P_B(x))$, with $P_A$ and $P_B$ defined as in \eqref{eqn:hard-thresholding} and \eqref{eqn:pseudo}, respectively, which resembles the ECME algorithm for known sparsity levels \cite{qiu2010}.

The projection $P_A$ \eqref{eqn:hard-thresholding}, known as hard thresholding, is an important part of many CS algorithms \cite{Blumensath2012,Blumensath2009,2010Blumensath,dai2009,qiu2010,qiu2011}. The projection $P_B$ \eqref{eqn:pseudo} also appears in the ECME algorithm \cite{qiu2010,qiu2011}. Normalized Iterative Hard Thresholding (NIHT) \cite{Blumensath2009} uses a calculation similar to $P_B$, replacing the pseudo-inverse with $\mu_t \Phi^T$ for an appropriately chosen scalar $\mu_t$.

Given that many algorithms consider the same types of projections as DM, any advantage achieved by DM must not come from the individual projections $P_A$ and $P_B$, but rather the way in which DM combines the two projections into a single iterative procedure. This is particularly true when comparing DM to the simple alternating map. Alternating between projections is guaranteed to find a point at the intersection of the two constraints if both are convex; however, if either of the constraints is not convex, it is easy for this scheme to get stuck in a local minimum that does not belong to intersection.

While many of the theoretical questions about the Difference Map remain open, it does come with a crucial guarantee here: even on nonconvex problems, a fixed point (meaning $D(x)=x$) implies that we have found a solution (meaning a point in $A \cap B$). 
To see this, note that $D(x) = x$ implies
\begin{equation}\label{eqn:DM2}
P_A \circ f_B(x) = P_B \circ f_A(x). 
\end{equation}
Thus we have found a point that exists in both $A$ and $B$. 
This leads us to believe that DM will find better sparse solutions when other algorithms are stuck in local minima.

Note that we are not the first to consider applying DM to compressed sensing. Qiu and Dogand\v{z}i\'{c} \cite{qiu2011} apply DM to the ECME algorithm (a variant of expectation maximization) in order to improve upon one of the two projections inside that algorithm. Although one of ECME's two projections uses DM \emph{internally}, ECME continues to combine the two projections in a simple alternating fashion. This is in stark contrast to our proposed algorithm, which uses DM \emph{externally} to the individual projections as a more intricate way of combining them. The resulting algorithm, called DM-ECME, is intended only for compressed sensing with non-negative signals. Because we consider different types of problems in this paper, we do not include DM-ECME in the comparisons below.

\subsection{Implementation Details}
We implemented the Difference Map in Matlab (code is available online\footnote{\url{http://web.cecs.pdx.edu/~landeckw/dm-cs0}}). All experiments were performed on  a computer with a 3 GHz quad-core Intel Xeon processor, running Matlab R2011a. We obtained Matlab implementations of LARS and StOMP from SparseLab v2.1 \cite{sparselab}. Implementations of AIHT \cite{Blumensath2012} and Subspace Pursuit \cite{dai2009} were found on the websites of the papers' authors. We also obtained Matlab implementations of ADMM \cite{Chartrand2013} and IRLS \cite{Chartrand2008} directly from the authors of the cited papers.

The implementations of LARS, Subspace Pursuit, AIHT, and StOMP are parameter-free. It was necessary to tune a single parameter for DM, and two parameters each for ADMM and IRLS. We tuned the parameters in two iterations of grid search. ADMM and IRLS required different parameters for the two different experiments presented in this paper (reconstruction from random measurements, and natural image reconstruction). Interestingly, DM performed well with the same parameter for both types of experiments.

We use ``training'' matrices of the same dimension, sparsity, and noise level as the ones presented in the figures of this paper in order to tune parameters. 
We chose parameters to minimize the MSE of the recovering $x$, averaged over all training problems. When tuning parameters for natural image reconstruction, we used a training set of 1000 image patches taken from the \emph{person} and \emph{hill} categories of ImageNet \cite{imagenet}, providing a good variety of natural scenery.

When tuning $\beta$ for DM, we first perform grid search with an interval of $0.1$, between $-1.2$ and $1.2$\footnote{The natural range for the parameter $\beta$ is [-1,1] (excluding 0), but Elser~\etal~\cite{Elser2007} report that occasionally values outside of this interval work well.
}. Next, in a radius of $0.5$ around the best $\beta$, we performed another grid search with an interval of $0.01$. Surprisingly, all $\beta$ in the interval $[-0.9, -0.1]$ appeared to be equally good for all problems reported in this paper. We chose $\beta = -0.14$ because it performed slightly better during our experiments, but the advantage over other $\beta \in [-0.9, -0.1]$ was not significant.

We use logarithmic grid search to tune the two parameters for ADMM and IRLS. First, we search parameter values by powers of ten, meaning $10^\alpha$, for $\alpha = -5, -4, \ldots, 5$. We then search in the neighborhood of best exponent $c$ by $\frac{1}{10}$ powers of ten, meaning $10^{c + \alpha}$ for $\alpha = -0.5, -0.4, \ldots, 0.5$.

For random measurements (the experiments in Section~\ref{sec:random-matrix}), this results in parameter values $\mu = 1.26\times 10^2, \lambda = 3.98\times 10^{-1}$ for ADMM, and $\alpha = 3.16\times 10^{-3}; \beta = 2.51\times 10^{-1}$. For natural image reconstruction (Section~\ref{sec:image}), we found $\mu = 1.58\times 10^2, \lambda = 1.0\times 10^{-1}$ for ADMM and $\alpha = 2.5\times 10^{-4}, \beta= 5\times 10^{-3}$ for IRLS. Note that the $\beta$ parameter for IRLS has nothing to do with the $\beta$ parameter for DM. We refer to both as $\beta$ only to remain consistent with the respective bodies of literature about each algorithm, but in what follows we will only refer to the parameter for DM. IRLS is capable of addressing the $\ell^p$ quasi-norm for a variety of values $0 < p \le 1$, while ADMM uses modifications of the $\ell^p$ quasi-norm designed to have a simple proximal mapping \cite{chartrand-2012-nonconvex}. In both cases we tried $p=\frac{1}{2}$ and $p=1$, and found $p=\frac{1}{2}$ to perform better.

%
%
%
%
%

\begin{figure*}[htb]
\centering
\includegraphics[height=4.5cm]{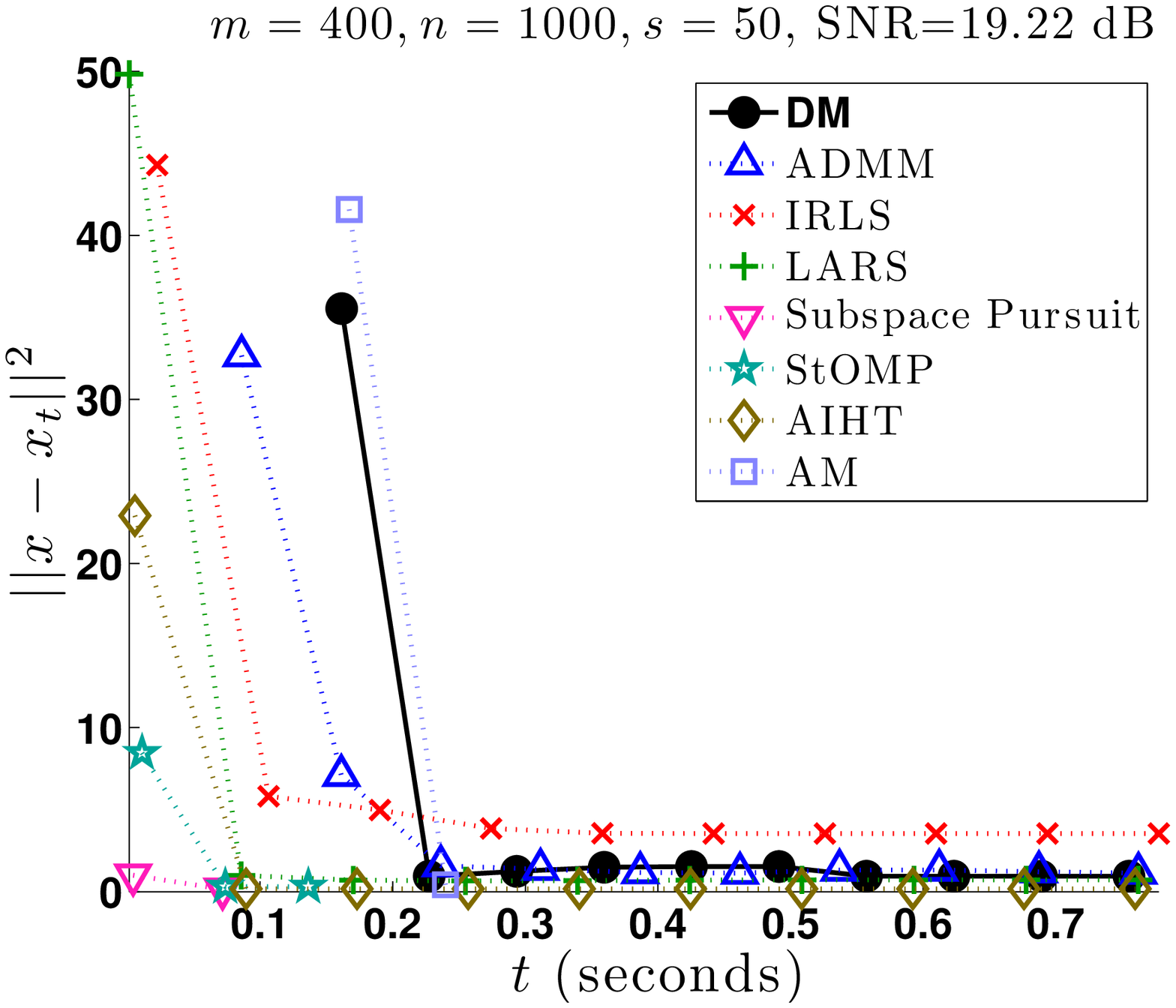}
\includegraphics[height=4.5cm]{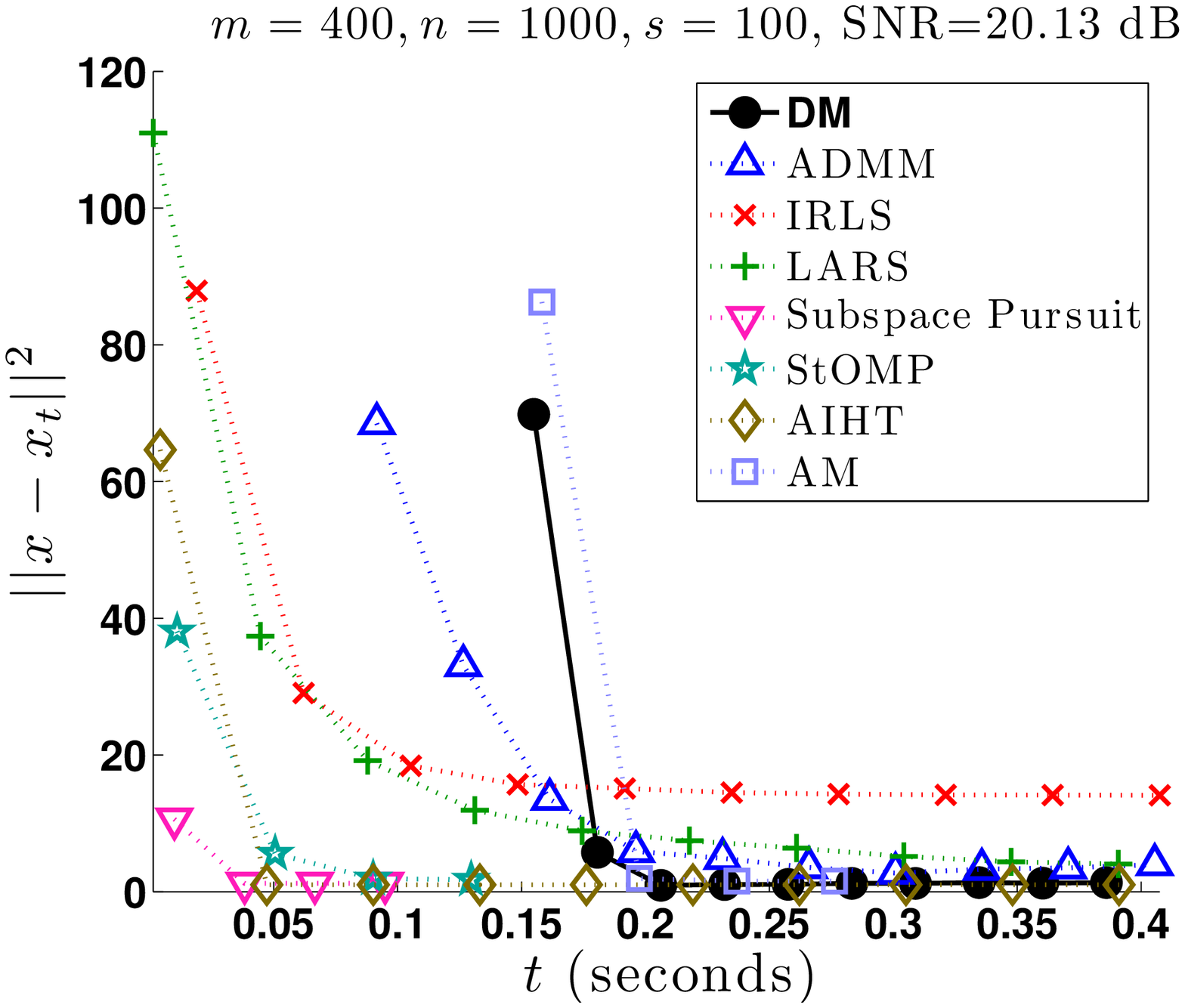}
\includegraphics[height=4.5cm]{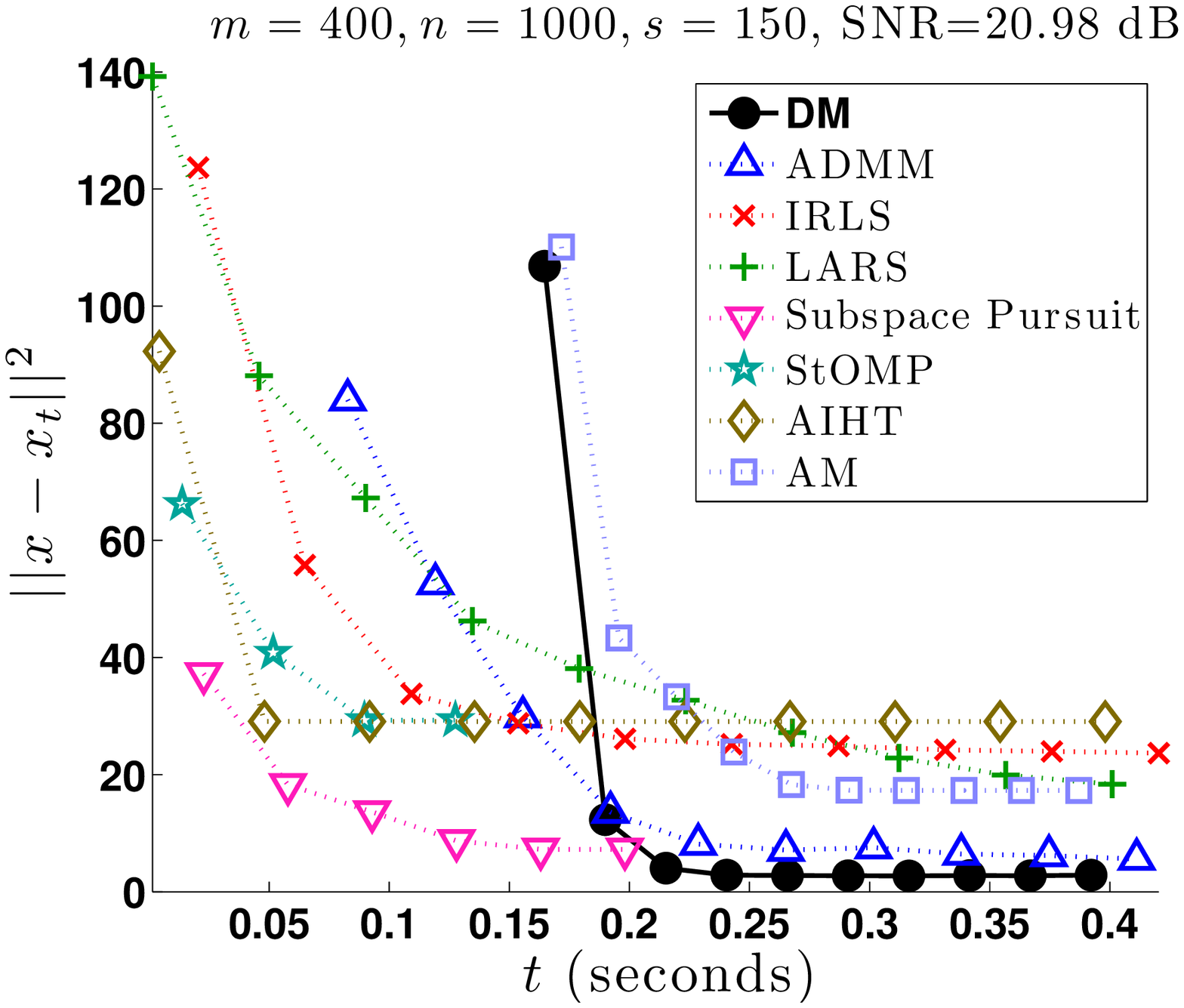}
\caption{Reconstructing signals with various levels of sparsity $s$. With sparser signals (left, middle), most algorithms get very close to the true signal. With less sparse signals (right), the Difference Map gets closer than other algorithms to recovering the signal. Each plot is averaged over ten runs, with $\epsilon$ chosen to give an SNR of approximately 20 dB, and $\Phi \in \R^{400\times1000}$.}
\label{fig:vary-s}
\end{figure*}

\begin{figure*}[htb]
\centering
\includegraphics[height=4.5cm]{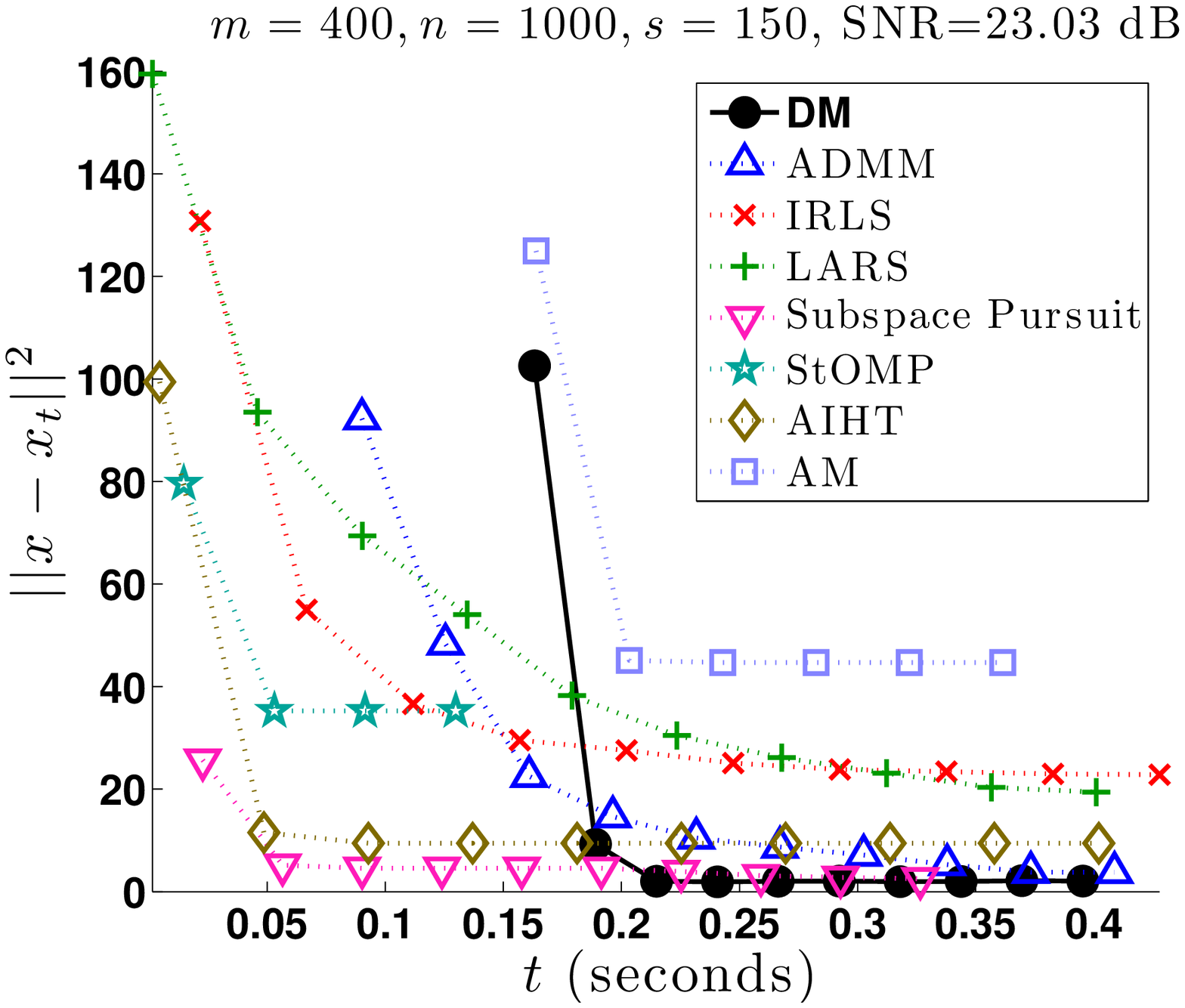}
\includegraphics[height=4.5cm]{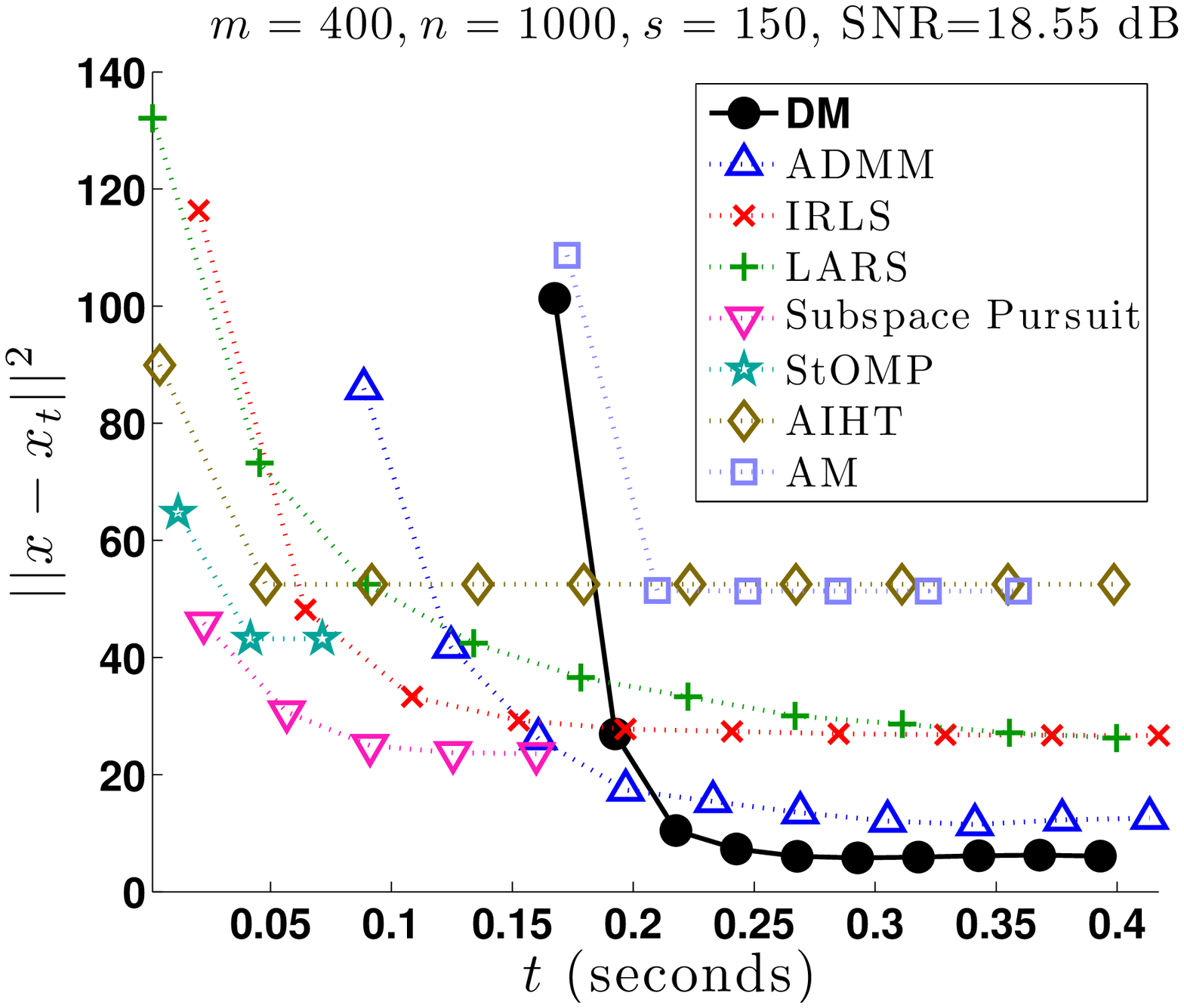}
\includegraphics[height=4.5cm]{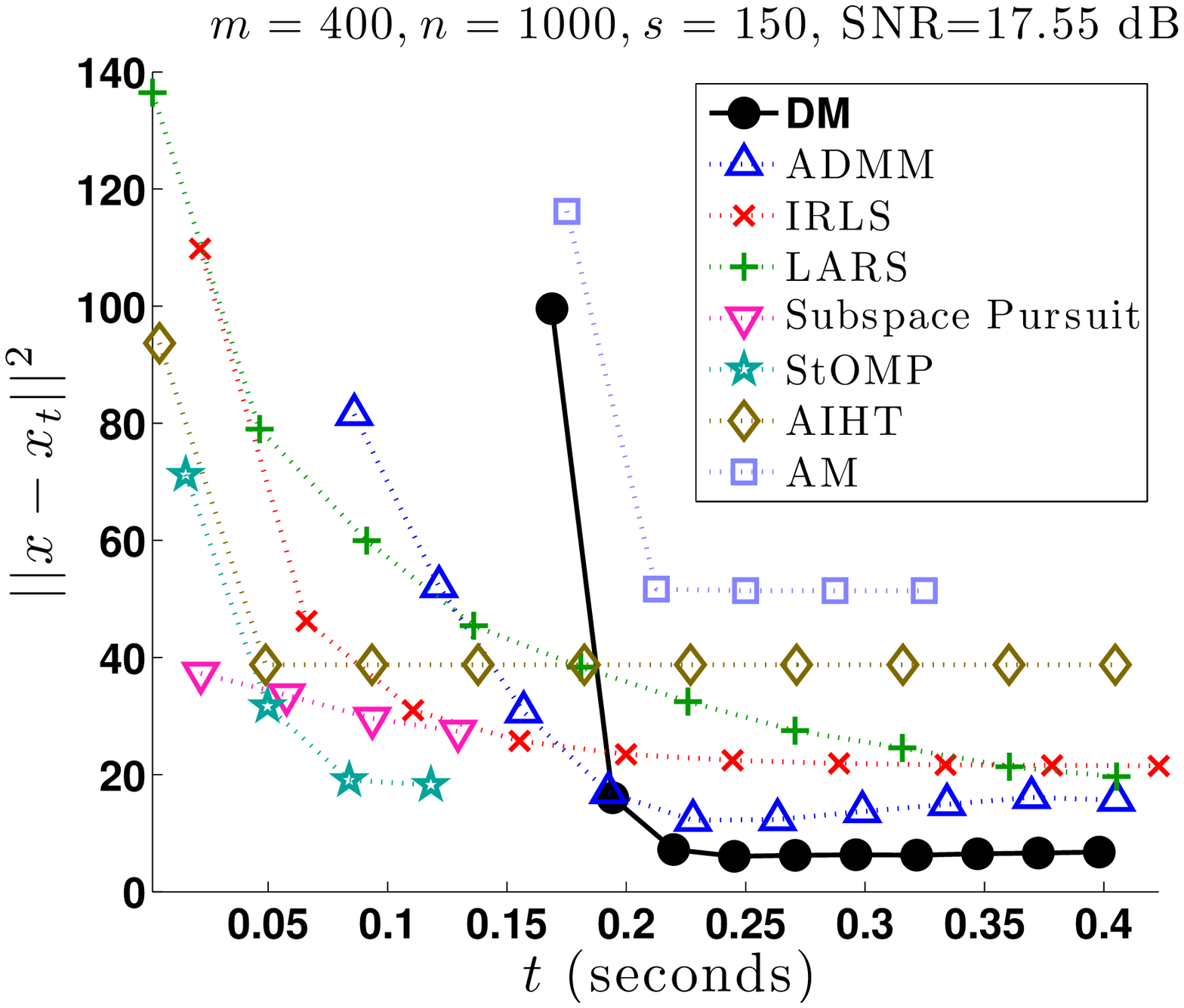}
\caption{Reconstructing signals with various levels of noise $\epsilon$. With less noise (left), most algorithms get very close to the true signal. With more noise (middle, right), the Difference Map gets closer than other algorithms to recovering the signal. Each plot is averaged over ten runs, with $s=150$ and $\Phi \in \R^{400\times1000}$.}
\label{fig:vary-noise}
\end{figure*}

\begin{figure*}[htb]
\centering
\includegraphics[height=4.5cm]{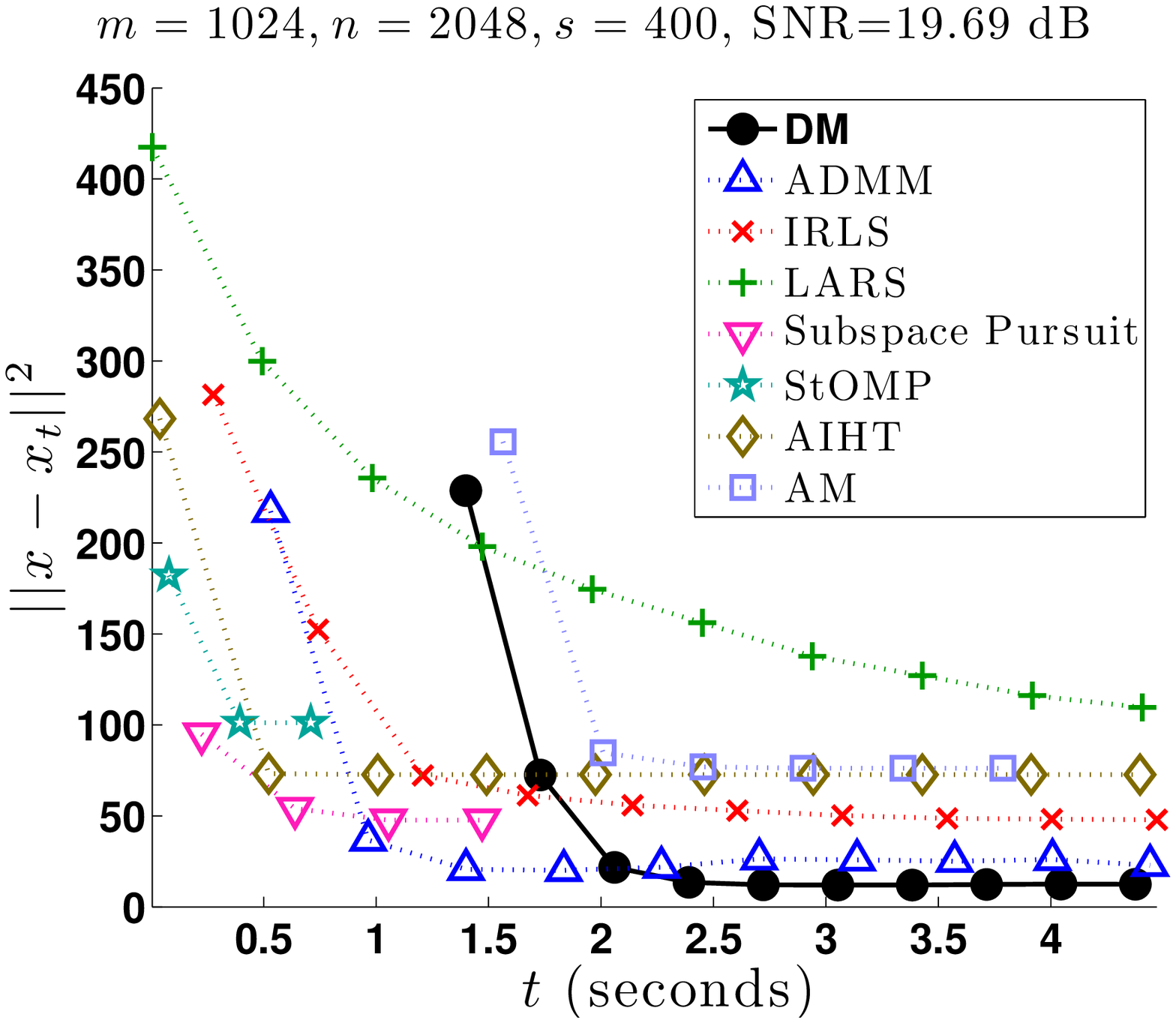}
\includegraphics[height=4.5cm]{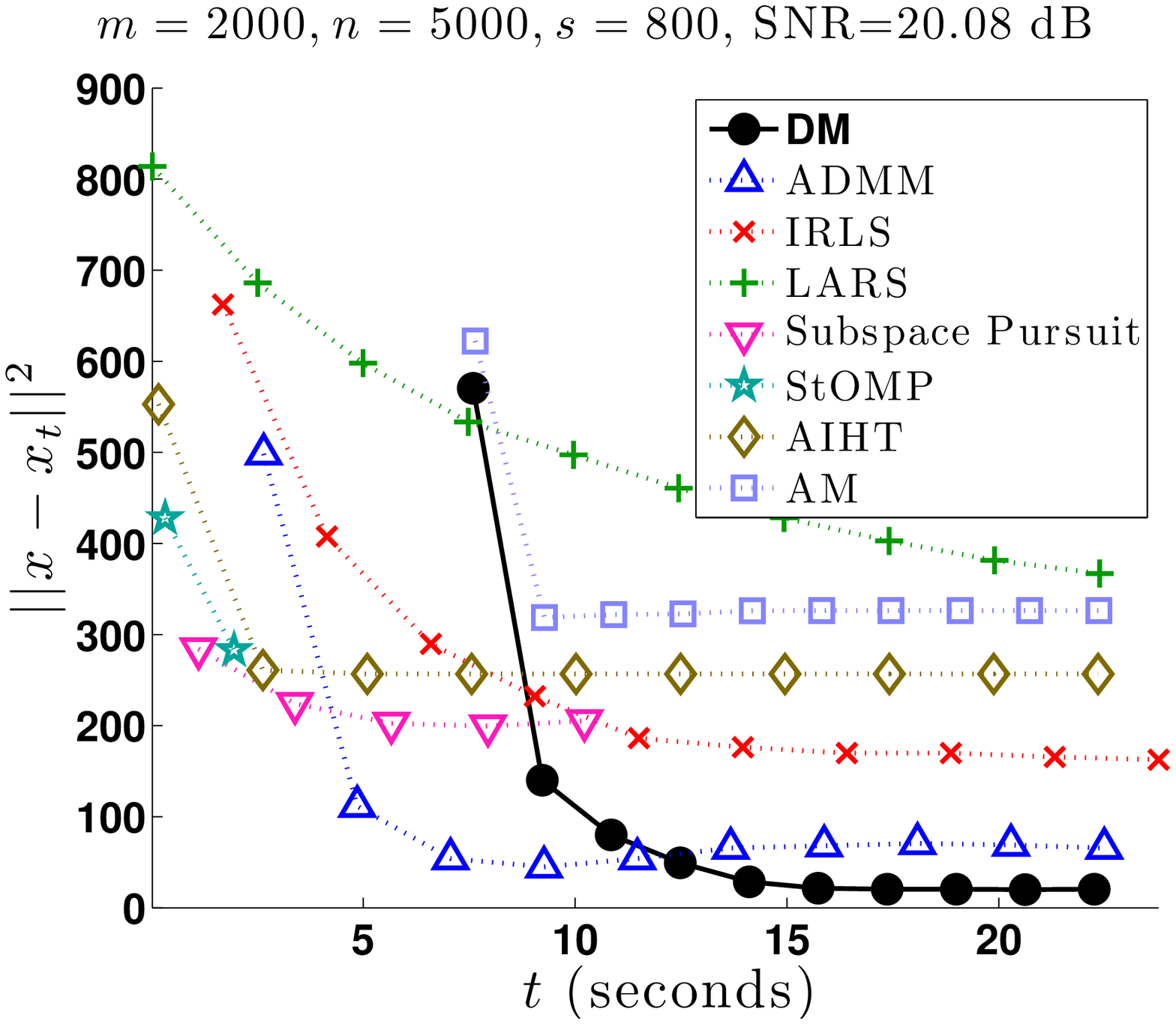}
\includegraphics[height=4.5cm]{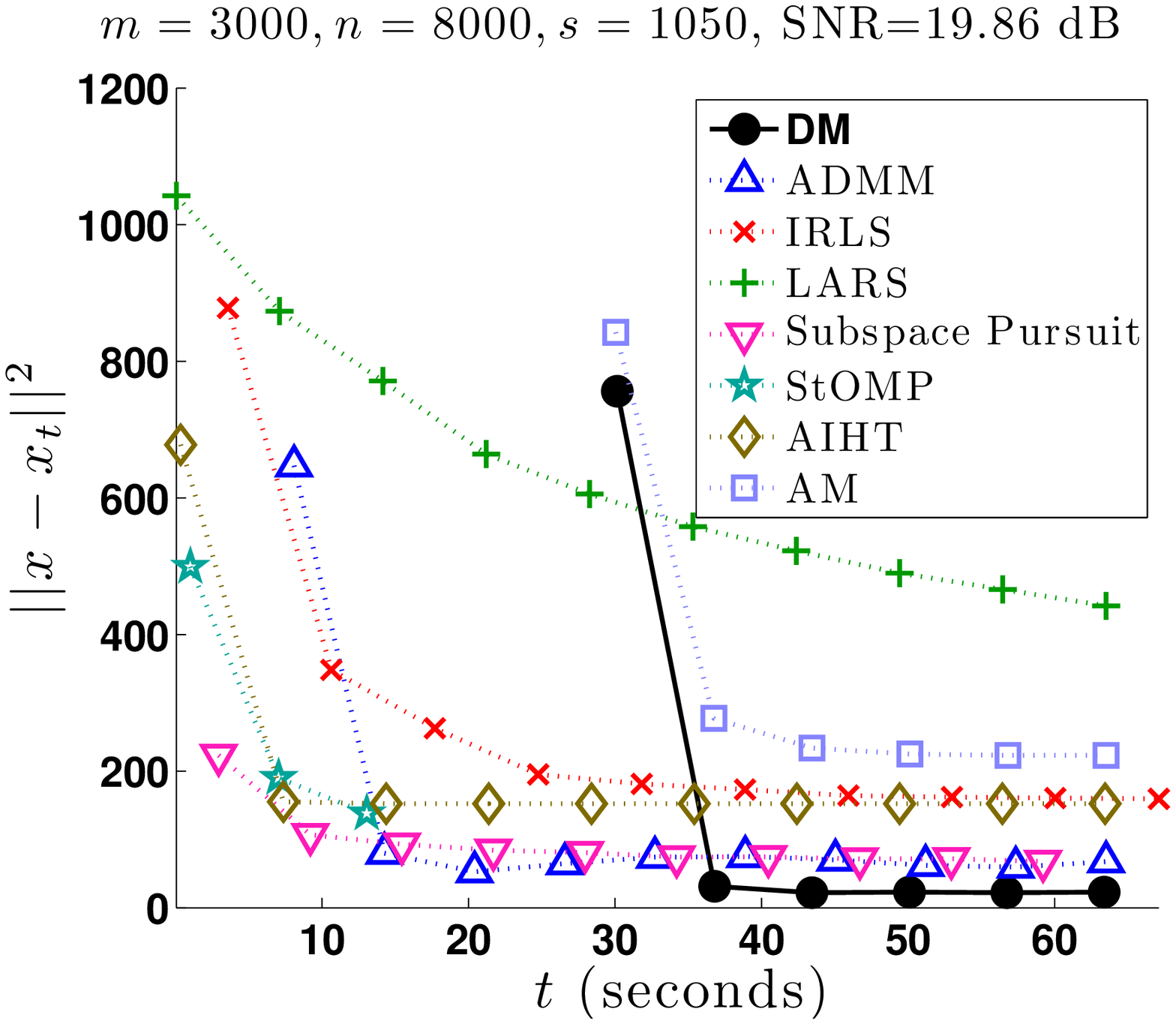}
\caption{In the low sparsity regime, the difference map outperforms other algorithms at recovering $x$ from a noisy observed signal with a wide variety of matrix sizes.  The sparsity ratio $s/n$ is approximately 1/3, and the noisy observation $\tilde{y} = \Phi x + \epsilon \cdot \mathcal{N}(0,1)$ has an SNR of approximately 20 dB. Each plot is averaged over ten runs.}
\label{fig:vary-m-n}
\end{figure*}

\section{Random Measurements} \label{sec:random-matrix}

In this Section, we compare the performance of DM to other algorithms when reconstructing signals from random measurements, testing a wide variety of matrix sizes, sparsity, and noise levels. Given positive integers $m, n,$ and $s$, we generate the random matrix $\Phi \in \R^{m \times n}$ with entries drawn from $\mathcal{N}(0,1)$. We then ensure that columns have zero mean and unit norm. We generate the $s$-sparse vector $x \in \R^n$ whose nonzero elements are drawn from $\mathcal{N}(0,1)$. We then calculate $y=\Phi x$, and the noisy ``observation'' $\tilde{y} = y + \epsilon \cdot \mathcal{N}(0,1)$. Finally, we ask each algorithm to reconstruct $x$ given only $\Phi$ and $\tilde{y}$. 

We measure runtime instead of iterations, as the time required per iteration varies widely for the algorithms considered. Additionally, the pre-computation for DM is the longest of any algorithm, requiring the pseudo-inverse of the dictionary. This pre-computation overhead is included in the timekeeping. 

In the first experiment, each algorithm attempts to reconstruct $x$ as we vary the sparsity level $s$. We choose $\epsilon$ so that the signal-to-noise ratio (SNR) is close to 20 dB. The results in Figure~\ref{fig:vary-s} demonstrate that for small values of $s$ (left, middle), meaning sparser signals, most algorithms are able to recover $x$ almost perfectly. As we increase the value of $s$ (right), meaning denser signals, other algorithms converge to undesirable minima. The Difference Map, however, continues to get very close to recovering $x$.

In the next experiment, each algorithm attempts to reconstruct $x$ as we vary the noise by changing $\epsilon$. $s$ is fixed at 150. The results in Figure \ref{fig:vary-noise} show that with less noise (left), meaning lower $\epsilon$ and higher SNR, several algorithms are able to get close to recovering the true signal $x$. As the noise increases (middle, right), meaning higher $\epsilon$ and lower SNR, only the Difference Map is able to get close to recovering the signal.

Note that DM and AM start ``late'' in all plots from Figures~\ref{fig:vary-s} and \ref{fig:vary-noise} because their pre-computation time is the longest (calculating $\Phi^+$). Despite using the same projections, we notice a large disparity in performance between DM and AM when $s=150$. Because the two algorithms both use the same two projections $P_A$ and $P_B$, this performance gap shows the power of combining two simple projections in a more elaborate way than simply alternating between them.

From the results in Figures~\ref{fig:vary-s} and \ref{fig:vary-noise}, we hypothesize that DM has a significant advantage with noisy, less sparse signals (higher $\epsilon$ and $s$) where other state-of-the-art compressed sensing algorithms get stuck in local minima or require a large amount of time to reach a good solution. We test this hypothesis with a variety of different matrix sizes in Figure \ref{fig:vary-m-n}, each time with a sparsity ratio $s/n$ of approximately $1/3$ and an SNR of approximately 20 dB. The results show that DM does indeed outperform other algorithms in this setting, for all matrix sizes tested.

%
%
%

\section{Sparse Coding Image Reconstruction}\label{sec:image}

The results from Section~\ref{sec:random-matrix} indicate that DM offers an advantage over other algorithms when the underlying signal is \emph{less} sparse and the observation is noisy. The less-sparse, noisy setting corresponds well to images which contain a large variety of textures, such as natural images. 
In this Section, we measure the sparse coding performance of the same algorithms as in the previous section (with the exception of AM, whose sparse coding performance was not competitive), by comparing reconstruction quality for a variety of images.

When reconstructing a large image, we  treat each $w \times w$ patch as an independent signal to reconstruct. Because our dictionary is constant, we only need to compute the pseudo-inverse in \eqref{eqn:pseudo} once. By amortizing the cost of the pseudo-inversion over all patches, 
this effectively allows DM to converge in less time per patch.
 We amortize the cost of pre-computation for other algorithms as well (most notably ADMM and IRLS).

In order to test the performance of the algorithms when reconstructing natural images, we require a dictionary learned for sparse image reconstruction. Dictionary learning is not the focus of this paper, but we present our method for completeness. The dictionary is trained with 10 million $20 \times 20$ image patches, and we choose to learn 1000 atoms, resulting in a dictionary of size $400 \times 1000$. We train the dictionary with patches from the \emph{person} and \emph{hills} category of ImageNet \cite{imagenet}, which provide a variety of natural scenery. We alternate sparse coding using 20 iterations of ADMM using $p$-shrinkage with $p = 1/2$ (see \cite{chartrand-2012-nonconvex} for details), with a dictionary update using the method of optimal directions \cite{engan-1999-method}.
We used ADMM as the sparse-coding algorithm simply because we had access to MPI-parallelized C code for this purpose. We do not believe that this gave an unfair advantage to ADMM, because the reconstructed images presented in this paper are separate from the dataset used to train the dictionary.

\begin{figure}[htb]
\centering
\includegraphics[width=6cm]{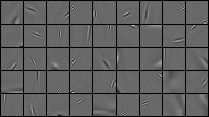}
\caption{
Example atoms from the dictionary $\Phi \in \mathbb{R}^{400\times1000}$ used for reconstruction. The dictionary contains elements of size $20\times20$, learned from 10 million image patches from the \emph{person} and \emph{hill} categories of ImageNet \cite{imagenet}.}
\label{fig:dict}
\end{figure}

Using 1,000 processors, the dictionary converged in about 2.5 hours. The training patches were reconstructed by the dictionary with an average of 29 nonzero components (out of 1000), and the reconstruction of the training images had a relative error of 5.7\%. The dictionary contains the typical combination of high- and low-frequency edges, at various orientations and scales. Some examples are shown in Figure~\ref{fig:dict}.

We reconstruct several natural images and measure the quality of the sparse reconstruction as a function of time. At time $t$, we measure the reconstruction quality of patch $y$ as follows. First, we perform hard-thresholding on the algorithm's current guess $x_t$, setting the $n-s$ smallest absolute values to zero, yielding the $s$-sparse vector $[ x_t ]_s$. We then calculate the reconstruction
\begin{equation*}
y_t = \Phi [ x_t ]_s
\end{equation*}
and measure the SNR of $y$ (the true image patch) to $y_t$. Thus we are measuring how well, at time $t$, the algorithm can create a \emph{sparse} reconstruction of $y$.
Note that algorithms returning a solution that is \emph{sparser} than required will not be affected by the hard-thresholding step. 

\begin{figure*}
\centering
\renewcommand{\tabcolsep}{1pt}
\begin{tabular}{cccccccc}
Original &
LARS & 
StOMP&  
IRLS &  
ADMM & 
Sub. Pursuit & 
AIHT &  
Diff. Map \\
 &  
13.93 dB & 
17.91 dB &  
20.57 dB &  
21.71 dB & 
16.97 dB &   
19.94 dB & 
 \textbf{23.73 dB}\\ 
\includegraphics[width=0.12\textwidth]{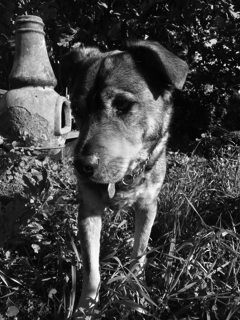}&
\includegraphics[width=0.12\textwidth]{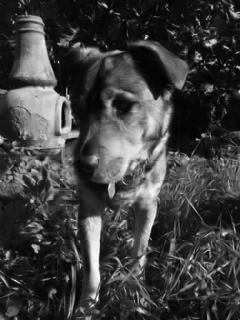}&
\includegraphics[width=0.12\textwidth]{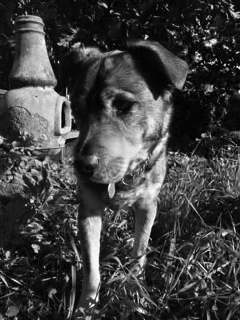}&
\includegraphics[width=0.12\textwidth]{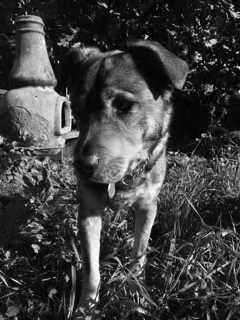}&
\includegraphics[width=0.12\textwidth]{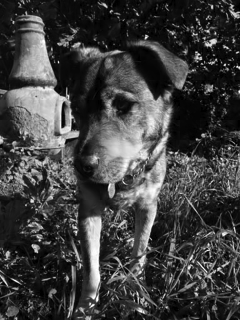}&
\includegraphics[width=0.12\textwidth]{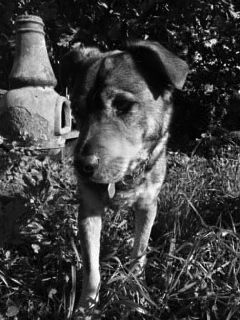}&
\includegraphics[width=0.12\textwidth]{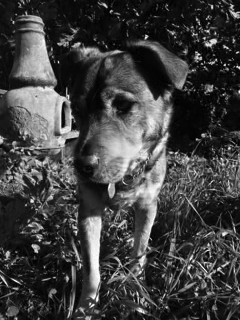}&
\includegraphics[width=0.12\textwidth]{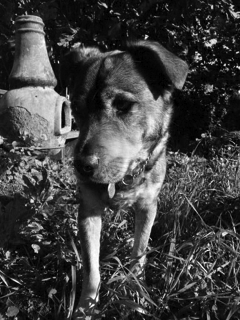} \\
\includegraphics[width=0.1\textwidth]{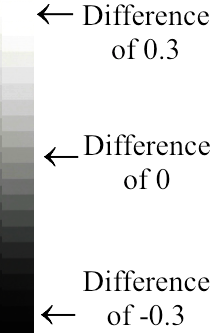}&
\includegraphics[width=0.12\textwidth]{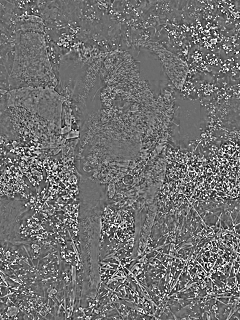}&
\includegraphics[width=0.12\textwidth]{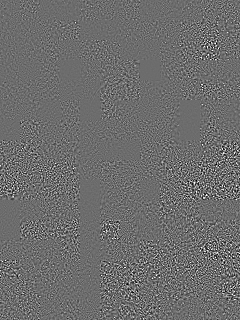}&
\includegraphics[width=0.12\textwidth]{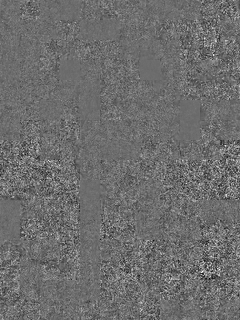}&
\includegraphics[width=0.12\textwidth]{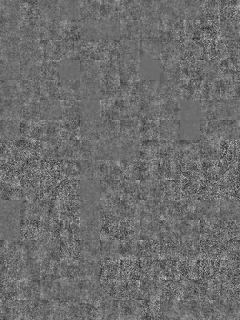}&
\includegraphics[width=0.12\textwidth]{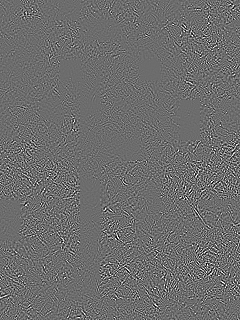}&
\includegraphics[width=0.12\textwidth]{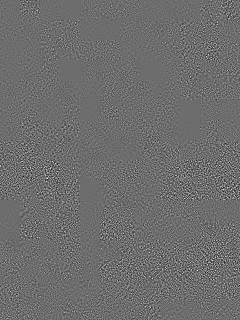}&
\includegraphics[width=0.12\textwidth]{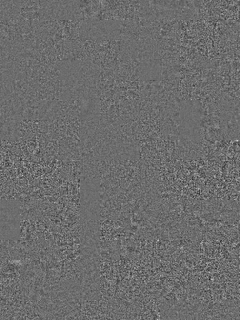} \\ \\ \\
\end{tabular}
\caption{Reconstructing a natural image. The Difference Map outperforms the other algorithms when reconstructing a $320 \times 240$ image of a dog (top). Difference images (bottom) show the difference between the reconstruction and the original image, which ranges from -0.3 (black) to 0.3 (white) -- original grayscale values are between 0 (black) and 1 (white). Results for $s = 200$ are shown, and the time limit was 0.1 seconds per $20 \times 20$ patch.} 
\label{fig:recon}
\end{figure*}

We reconstruct a $320 \times 240$ image of a dog, seen in Figure~\ref{fig:recon}, using the $400 \times 1000$ dictionary from Figure~\ref{fig:dict}. 
We measure results for both $s=100$ and $s=200$, as well as $t=0.05$ and $t=0.1$\footnote{
We measure time in seconds per $20 \times 20$ patch. Thus $t=0.05$ corresponds to approximately 10 seconds for the entire image, and $t=1$ to 20 seconds for the entire images. 
The astute reader will notice that when $\Phi$ has dimension $400 \times 1000$, it takes longer than 0.1 seconds to compute $\Phi^+$. This can be seen in Figures~\ref{fig:vary-s} and \ref{fig:vary-noise}, where it takes almost 0.2 seconds for DM to finish calculating $\Phi^+$ and begin searching for $x$. However, because we only need to calculate the pseudo-inverse once for the entire image, this start-up cost is amortized over all patches and becomes negligible.}. 
The results in Table~\ref{table:dog} show that DM consistently achieves the highest SNR. Furthermore, while increasing $s$ and $t$ tend to improve each algorithm's reconstruction, the relative quality between algorithms stays fairly consistent.

\begin{table}[h]
\caption{Signal to noise ratio (SNR, in decibels) of the reconstructed image from Figure~\ref{fig:recon}. We test various sparsity levels $s$ and various runtimes (seconds per $20 \times 20$ patch). The difference map consistently achieves the highest SNR.}
\begin{center}
\begin{tabular}{|r|c|c|c|c|}\hline
&\small{$s=100$}&\small{$s=100$}&\small{$s=200$}&\small{$s=200$}\\
& \small{$t=0.05$} & \small{$t=0.1$} & \small{$t=0.05$} &  \small{$t=0.1$} \\ \hline
Diff. Map & \textbf{17.02} &\textbf{17.55} & \textbf{22.44} & \textbf{23.73} \\ \hline
ADMM & 15.27 & 16.59 & 19.37&21.71\\ \hline
IRLS& 13.39 & 14.87 & 18.14 & 20.57\\ \hline
Sub. Pursuit& 16.58 & 16.67 & 16.99 & 16.97\\ \hline
LARS& 11.35 & 13.88 & 11.30 & 13.93\\ \hline
AIHT& 14.78 & 15.70 & 18.79 & 19.94\\ \hline
StOMP & 15.40 & 15.39 & 17.84 & 17.91 \\ \hline
\end{tabular}
\end{center}
\label{table:dog}
\end{table}

The highest quality reconstructions, achieved with $s=200$ and $t=0.1$, are shown in Figure~\ref{fig:recon} (top row). While some algorithms fail to reconstruct details in the animal's fur and the grass, many algorithms reconstruct the image well enough to make it difficult to find errors by mere visual inspection. We show the difference between the reconstructions and the original image (Figure~\ref{fig:recon}, bottom row), where a neutral gray color in the difference image corresponds to a perfect reconstruction of that pixel; white and black are scaled to a difference of 0.3 and -0.3, respectively (the original image was scaled to the interval [0,1]).

\begin{figure}[h]
\begin{center}
\begin{tabular}{|c|c|c|} \hline
\multirow{7}{*}{\includegraphics[height=2.95cm]{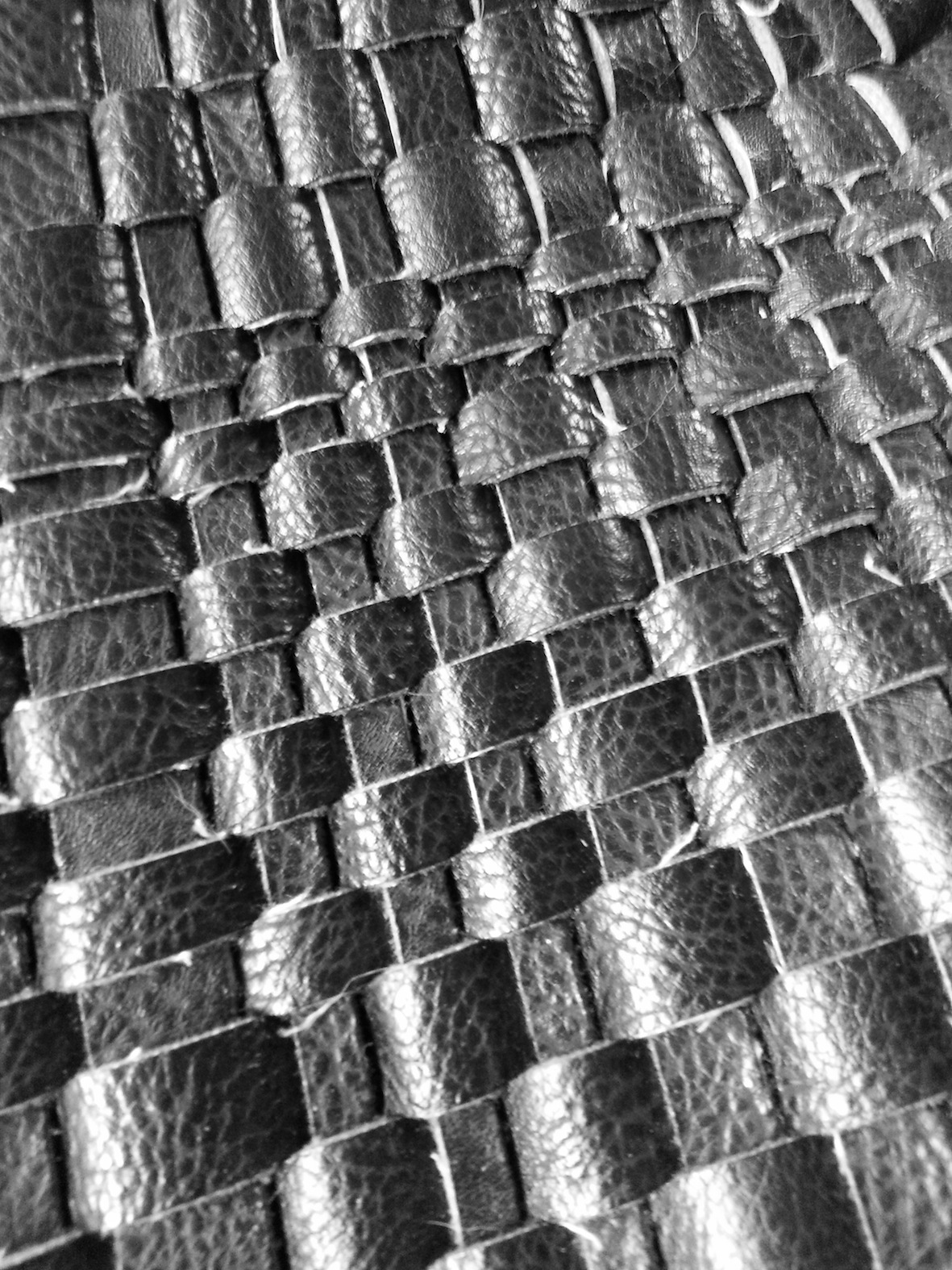}} 
& \textbf{Difference Map} & \textbf{23.33 } \\ \cline{2-3}
& ADMM & 20.80  \\  \cline{2-3}
& IRLS & 20.50  \\  \cline{2-3}
& LARS & 13.62  \\  \cline{2-3}
& Subspace Pursuit & 17.96  \\  \cline{2-3}
& StOMP & 19.48  \\  \cline{2-3}
& AIHT & 19.41  \\  \hline \hline
\multirow{7}{*}{
	\includegraphics[height=2.95cm]{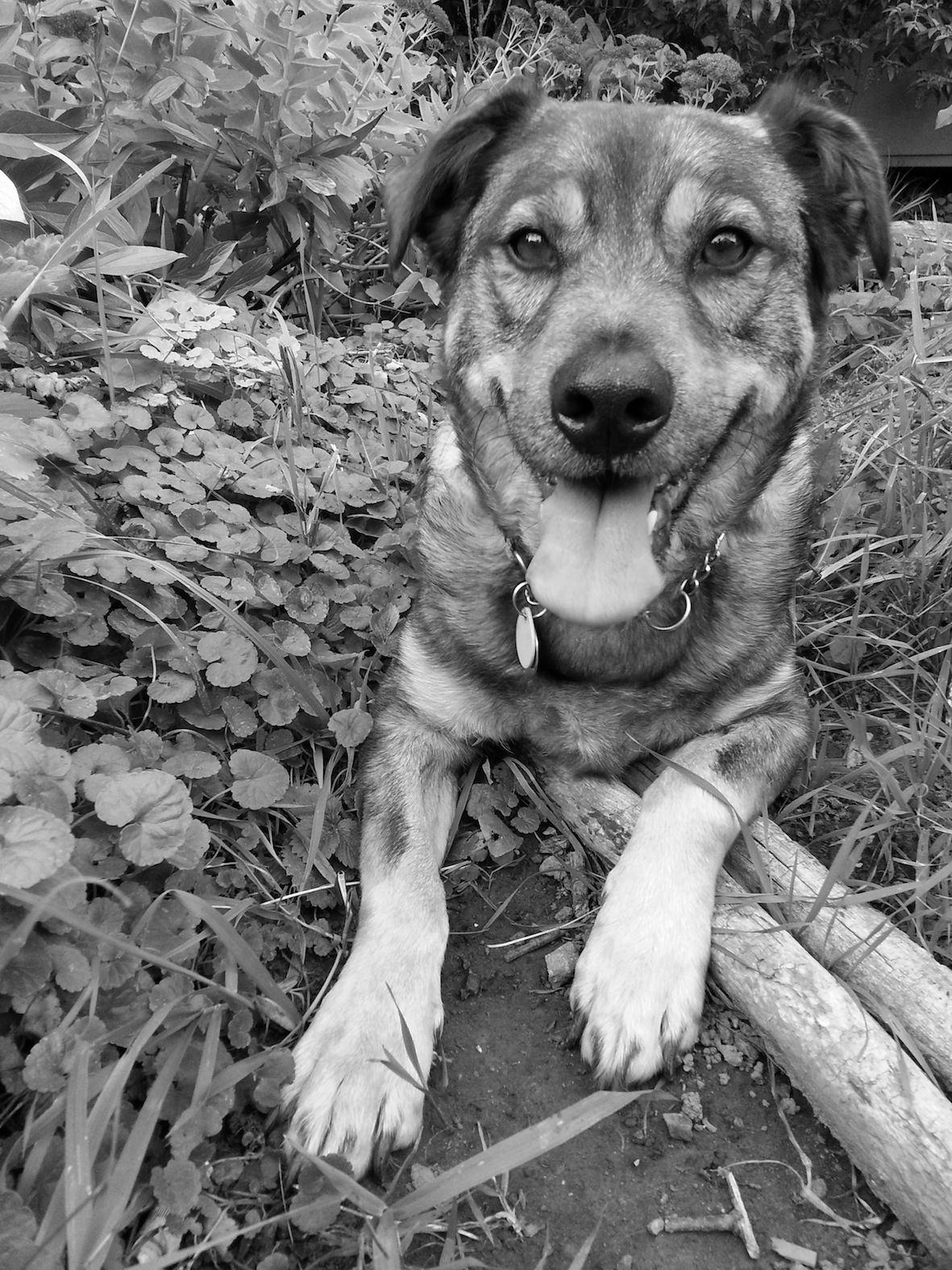}
} 
& \textbf{Difference Map} & \textbf{23.06 } \\ \cline{2-3}
& ADMM & 20.67  \\  \cline{2-3}
& IRLS & 20.59  \\  \cline{2-3}
& LARS & 14.37  \\  \cline{2-3}
& Subspace Pursuit & 17.44  \\  \cline{2-3}
& StOMP & 18.78  \\  \cline{2-3}
& AIHT & 19.44  \\  \hline \hline
\multirow{7}{*}{
	\includegraphics[width=3cm]{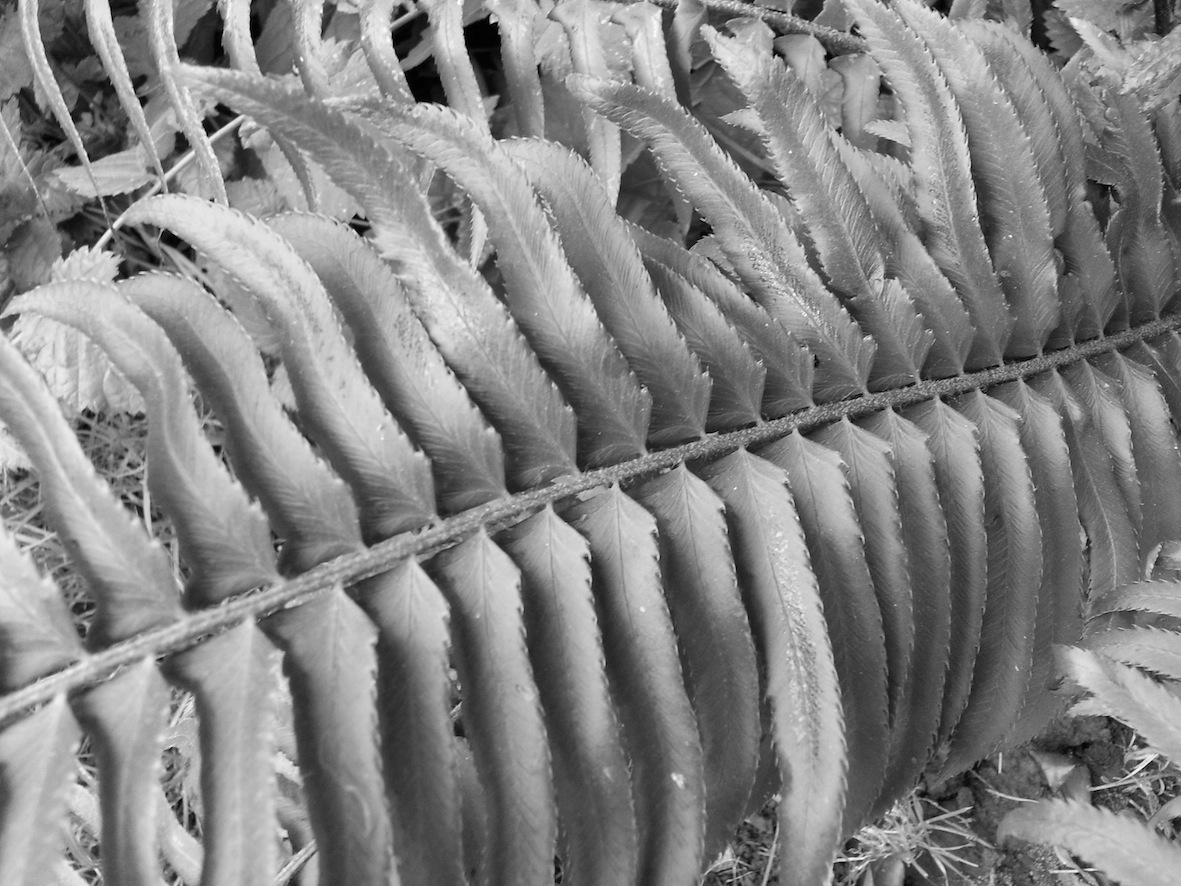}
} 
& \textbf{Difference Map} & \textbf{25.39 } \\ \cline{2-3}
& ADMM & 22.72  \\  \cline{2-3}
& IRLS & 23.55  \\  \cline{2-3}
& LARS & 17.60  \\  \cline{2-3}
& Subspace Pursuit & 20.66  \\  \cline{2-3}
& StOMP & 23.37  \\  \cline{2-3}
& AIHT & 21.98  \\  \hline \hline
\multirow{7}{*}{
	\includegraphics[width=3cm]{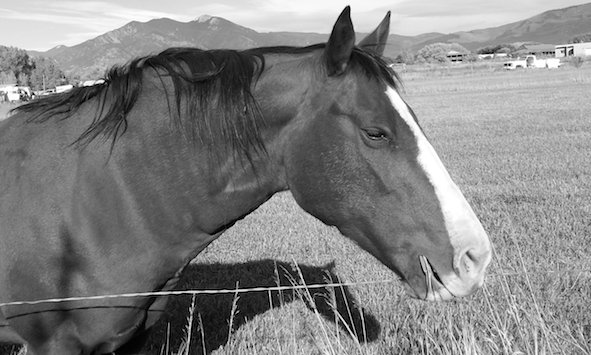}
} 
& \textbf{Difference Map} &\textbf{23.95 } \\ \cline{2-3}
& ADMM & 21.90  \\  \cline{2-3}
& IRLS & 20.90  \\  \cline{2-3}
& LARS & 14.27  \\  \cline{2-3}
& Subspace Pursuit & 17.14  \\  \cline{2-3}
& StOMP & 17.97  \\  \cline{2-3}
& AIHT & 20.34  \\  \hline 
\end{tabular}
\caption{The Difference Map regularly outperforms other algorithms in finding sparse reconstructions of a variety of images. We measure the SNR in decibels (right column) between the reconstruction and the original image (left column). Images are scaled to $240 \times 320$ pixels ($320 \times 240$ for horizontal images). Reconstructions have sparsity $s = 200$, and are completed in 20 seconds per image (approximately 0.1 seconds per $20 \times 20$ patch). The dictionary $\Phi$ is the same as in Figure~\ref{fig:dict}.}
\label{fig:many-images}
\end{center}
\end{figure}

The advantage of DM over other algorithms, when sparsely reconstructing images, can be seen with a large variety of images. In Figure~\ref{fig:many-images}, we see that DM consistently achieves the best reconstruction. All original images are available online\footnote{\url{http://web.cecs.pdx.edu/\~landeckw/dm-cs0}}.

\section{Conclusions}\label{sec:conclusions}
We have presented the Difference Map, a method of finding a point at the intersection of two constraint sets, and we have introduced an implementation of DM for compressed sensing and sparse coding. The constraint-set formulation is a natural fit for sparse recovery problems, in which we have two competing constraints for $x$: to be consistent with the data $y$, and to be sparse.

When the solution $x$ is very sparse and the observation $\tilde{y}$ is not too noisy, DM takes more time in finding the same solution as competing algorithms. However, when the solution $x$ is \emph{less} sparse and when the observation $\tilde{y}$ is noisy, DM outperforms state of the art sparse recovery algorithms. The noisy, less sparse setting corresponds well to reconstructing natural images, which can often require a large number of components in order to accurately reconstruct. Experiments show that DM performs favorably in reconstructing a variety of images, with a variety of parameter settings.

Parameter tuning can present a laborious hurdle to the researcher. DM requires tuning only a single parameter $\beta$. For all experiments in this paper (natural image reconstruction for various images; reconstruction with random matrix dictionaries of various sizes, with varying amounts of sparsity and noise), we found DM to work almost equally as well for all $-0.9 \le \beta \le -0.1$. The robustness of DM under such a wide variety of parameter values and problems makes DM a very competitive choice for compressed sensing.


The robustness of DM comes from how it combines two simple projections into a single iterative procedure. The Alternating Map (AM) combines the same projections in a simple alternating fashion, and struggles in almost all experiments. The gap in performance between these two methods demonstrates the power of combining multiple constraints in a more perspicacious way.

Finally, we recall that performance in all experiments was measured as a function of time, which would seem to put DM at a natural disadvantage to other algorithms: DM requires the pseudo-inverse of the dictionary, computing which requires more time than any other algorithm's pre-computation. Despite this, DM consistently outperforms other algorithms.

{\small
\bibliographystyle{ieee}
\bibliography{DM}
}

\end{document}